\documentclass[sigconf]{acmart}

\usepackage{multirow}
\usepackage{makecell}

\AtBeginDocument{%
  }

\setcopyright{acmlicensed}
\copyrightyear{2025}
\acmYear{2025}
\setcopyright{acmlicensed}
\acmConference[KDD '25] {Proceedings of the 31st ACM SIGKDD Conference on Knowledge Discovery and Data Mining V.2}{August 3--7, 2025}{Toronto, ON, Canada.}
\acmBooktitle{Proceedings of the 31st ACM SIGKDD Conference on Knowledge Discovery and Data Mining V.2 (KDD '25), August 3--7, 2025, Toronto, ON, Canada}
\acmISBN{979-8-4007-1454-2/25/08}
\acmDOI{10.1145/3711896.3737238}

\begin{document}

\title{LaDEEP: A Deep Learning-based Surrogate Model for Large Deformation of Elastic-Plastic Solids}


\author{Shilong Tao}
\orcid{0009-0003-2497-3413}
\affiliation{%
  \institution{Peking University}
  \department{School of Computer Science}
  \city{Beijing}
  \country{China}
}
\email{shilongtao@stu.pku.edu.cn}

\author{Zhe Feng}
\orcid{0009-0000-8682-3769}
\affiliation{%
  \institution{Peking University}
  \department{School of Computer Science}
  \city{Beijing}
  \country{China}
}
\email{zhe.feng27@stu.pku.edu.cn}

\author{Haonan Sun}
\orcid{0009-0007-8784-6963}
\affiliation{%
  \institution{Peking University}
  \department{School of Computer Science}
  \city{Beijing}
  \country{China}
}
\email{sunhaonan331@pku.edu.cn}

\author{Zhanxing Zhu}
\orcid{0000-0002-2141-6553}
\affiliation{%
  \institution{University of Southampton}
  \department{School of Electrical and Computer Science}
  \city{Southampton}
  \country{United Kingdom}
}
\email{z.zhu@soton.ac.uk}

\author{Yunhuai Liu}
\authornote{Corresponding author.}
\orcid{0000-0002-1180-8078}
\affiliation{%
  \institution{Peking University}
  \department{School of Computer Science}
  \city{Beijing}
  \country{China}
}
\email{yunhuai.liu@pku.edu.cn}

\renewcommand{\shortauthors}{Shilong Tao, Zhe Feng, Haonan Sun, Zhanxing Zhu, and Yunhuai Liu.}

\begin{abstract}
Scientific computing for large deformation of elastic-plastic solids is critical for numerous real-world applications. Classical numerical solvers rely primarily on local discrete linear approximation and are constrained by an inherent trade-off between accuracy and efficiency. Recently, deep learning models have achieved impressive progress in solving the continuum mechanism. While previous models have explored various architectures and constructed coefficient-solution mappings, they are designed for general instances without considering specific problem properties and hard to accurately handle with complex elastic-plastic solids involving contact, loading and unloading. In this work, we take stretch bending, a popular metal fabrication technique, as our case study and introduce LaDEEP, a deep learning-based surrogate model for \textbf{La}rge \textbf{De}formation of \textbf{E}lastic-\textbf{P}lastic Solids. We encode the partitioned regions of the involved slender solids into a token sequence to maintain their essential order property. To characterize the physical process of the solid deformation, a two-stage Transformer-based module is designed to predict the deformation with the sequence of tokens as input. Empirically, LaDEEP achieves five magnitudes faster speed than finite element methods with a comparable accuracy, and gains 20.47\% relative improvement on average compared to other deep learning baselines. We have also deployed our model into a real-world industrial production system, and it has shown remarkable performance in both accuracy and efficiency. Code is available at \href{https://github.com/therontau0054/LaDEEP}{https://github.com/therontau0054/LaDEEP}. 
\end{abstract}



\begin{teaserfigure}
  \centering   
  \includegraphics[width=0.9\linewidth,keepaspectratio]{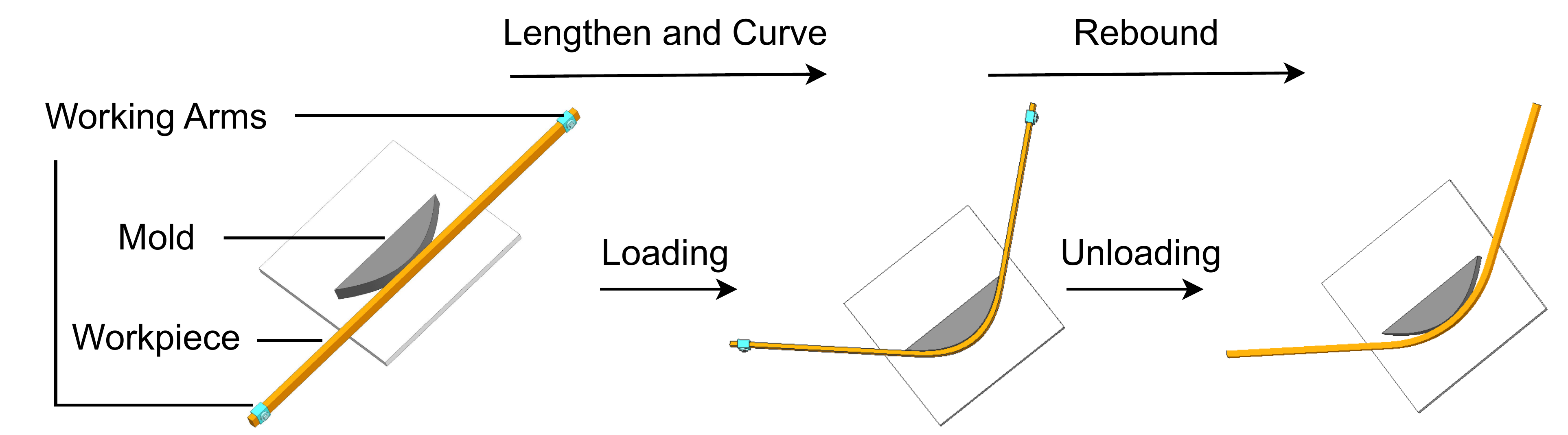}
    \caption{The stretch-bending process. It consists of two elastic-plastic stages including loading process and unloading process.}
\label{figure2}
\end{teaserfigure}


\maketitle

\section{Introduction}
\label{Introduction}

The scientific computing for large deformation of elastic-plastic solids \cite{bathe1976elastic} is essential in continuum mechanics, which is widely used in various areas such as civil engineering \cite{abourizk1998framework}, aerospace \cite{phanden2021review}, and nuclear materials \cite{allen2012comprehensive}. The deformation of a solid typically results from the application of external loads and contact constraints \cite{shabana2020dynamics, wriggers2006computational, bathe2006finite}. Large deformation occurs when the deformation extent becomes significant enough to invalidate the assumptions of infinitesimal strain theory \cite{bower2009applied}. Metals are particularly important subjects of study due to their distinct elastic and plastic behaviors. In the elastic regime, metals tend to return to their original shape after deformation, but only up to a certain threshold. Beyond this limit, when sufficient load is applied and the material enters the plastic state, permanent deformation occurs. Upon the release of the load, the elastic component of the deformation will recover, while the plastic deformation remains, causing the material to rebound slightly and assume its final shape. An accurate and efficient solver for such complex task is in urgent demand.

Figure~\ref{figure2} illustrates a practical example known as stretch bending \cite{clausen2000stretch}, one of the most widely used metal fabrication techniques
\cite{murr2012metal}. This process consists of two consecutive stages: loading and unloading. During the loading phase, the metal workpiece is positioned on the machine and securely held by two working arms. These arms move and rotate, applying force to the workpiece, causing it to lengthen and curve around a mold, which acts as a contact constraint. In the unloading phase, the applied loads are released, allowing the metal to rebound and assume its final shape. Our primary interest lies in predicting the final shape of the workpiece given the influence of the applied loads and contact constraints.

\begin{figure}[t]
    \centering
    \includegraphics[width=0.95\linewidth,keepaspectratio]
    {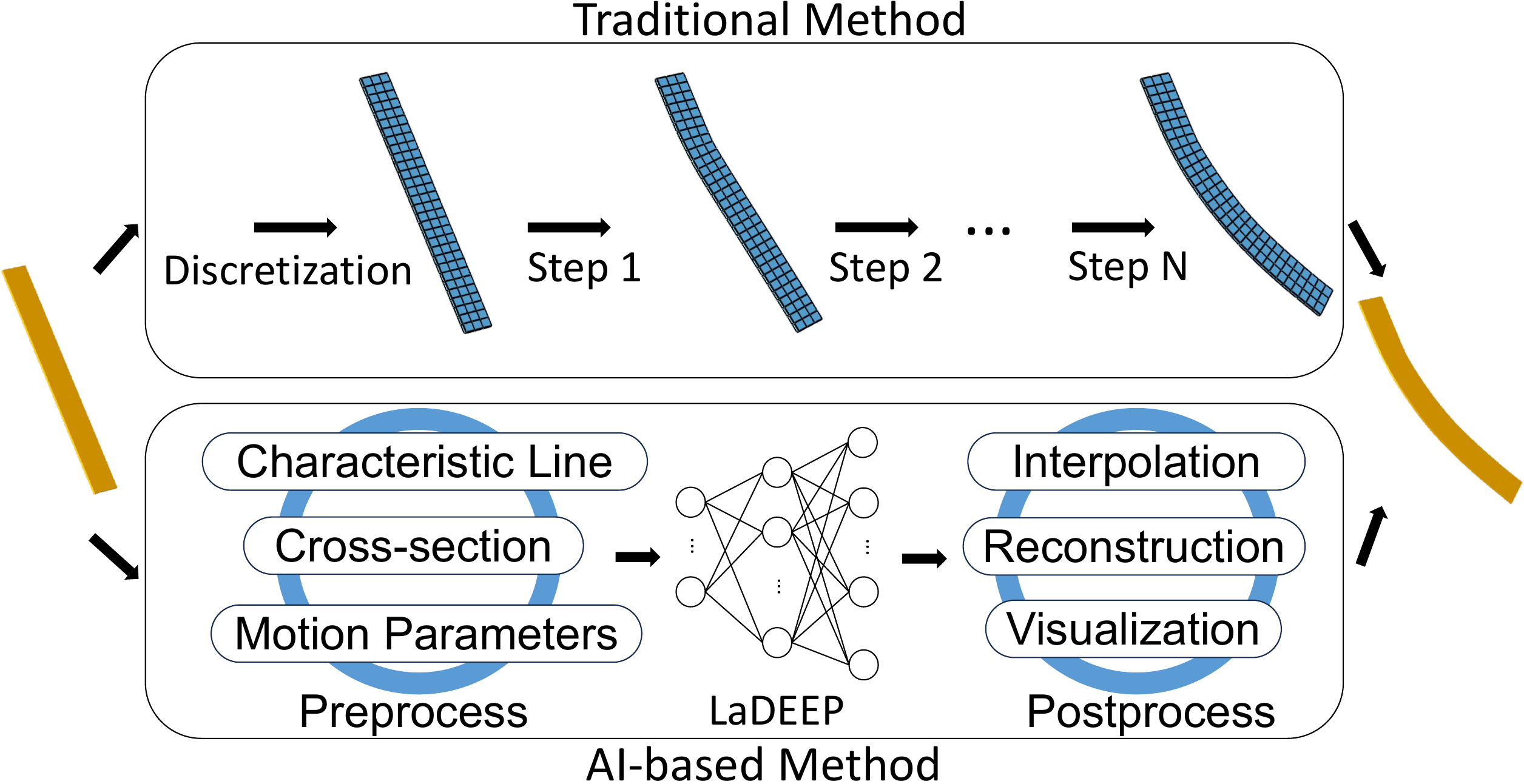}
    \caption{Traditional Method and AI-based Method.}
    \label{figure1:a}
    \vspace{-5pt}
\end{figure}

This fabrication technique involves large deformation of elastic-plastic solids, governed by partial differential equations (PDEs). The workpiece corresponds to the solution domain, the mold represents the boundary condition, and the loads from the movement of the working arms are modeled as source terms (external forces). Traditional methods, such as Finite Element Methods (FEM) \cite{reddy2019introduction}, depicted in Figure~\ref{figure1:a}, rely on local discrete linear approximation. The material is subdivided into smaller elements through meshing techniques in the solution domain \cite{alliez2005variational}, and solutions are incrementally approximated. However, these methods face a trade-off between computational accuracy and efficiency due to the discretization. Higher accuracy could be achieved with finer and more regular elements, though in practical cases, a large number of irregularly shaped elements may be encountered. Moreover, discretizing continuous domains introduces tens of thousands of degrees of freedom, resulting in an exponential rise in computational complexity \cite{koppen2000curse}.

Recently, deep learning (DL) has demonstrated significant potential in solving PDEs \cite{raissi2019physics, lu2019deeponet, li2022fourier, wen2022u}. Physics-Informed Neural Networks (PINNs) \cite{raissi2019physics} incorporate PDE constraints into the loss function and leverage automatic differentiation to optimize the model, effectively transforming the network into a solver for specific PDE instances. However, PINNs struggle to generalize to similar instances without retraining. Neural operators, like FNOs \cite{li2022fourier, wen2022u, rahman2022u, gupta2021multiwavelet, li2020neural, li2023fourier} and DeepONet \cite{lu2019deeponet}, learn the mappings between infinite-dimensional function spaces, offering a broader applicability. Transformer-based \cite{vaswani2017attention} methods have also achieved noteworthy progress in PDE-solving \cite{liu2021swin, li2024scalable, hao2023gnotgeneralneuraloperator, xiao2024improvedoperatorlearningorthogonal}. LSM \cite{wu2023solving} employs an attention-based hierarchical projection network that incorporates spectral methods \cite{gottlieb1977numerical} to reduce the high-dimensional data into a compact latent space in linear time. Similarly, Transolver \cite{wu2024transolver} introduces physics-attention to adaptively partition the discretized domain into a series of learnable slices, capturing the underlying physical states. While these methods are generally effective for problems in fluid mechanics or simple solid mechanics, they tend to underperform in our problem, which involves non-smooth boundary conditions (solid contacts) and multi-stage processes (unloading and loading) due to the ignorance of characteristics of large deformation and staged modeling.

To address these challenges, we begin by analyzing the inherent properties of the task. Focusing on a highly relevant industrial scenario, the stretch bending, we introduce LaDEEP as a surrogate model to approximate the complex multi-stage process. The model follows an Encoder-Processor-Decoder procedure~\cite{battaglia2018relational}. We develop several specialized modules to encode property-aware token sequences, which represent sequentially partitioned regions corresponding to slender solids that retain essential order properties. Then, a two-stage Transformer-based module, the Deformation Predictor (DP), is proposed as a processor to characterize the loading and unloading process. By applying attention mechanisms to the  token sequences , DP effectively captures the complex underlying interactions between objects. To validate our approach, we create a new dataset to fill the gap of data scarcity in this field and conduct extensive experiments comparing LaDEEP with classical FEM method and other deep learning models. LaDEEP demonstrates significant performance improvement over other alternatives. More importantly, we have successfully deployed LaDEEP in practical scenario. Our main contributions are as follows.

\begin{figure*}[t]
  \centering
  \includegraphics[width=\linewidth,keepaspectratio]{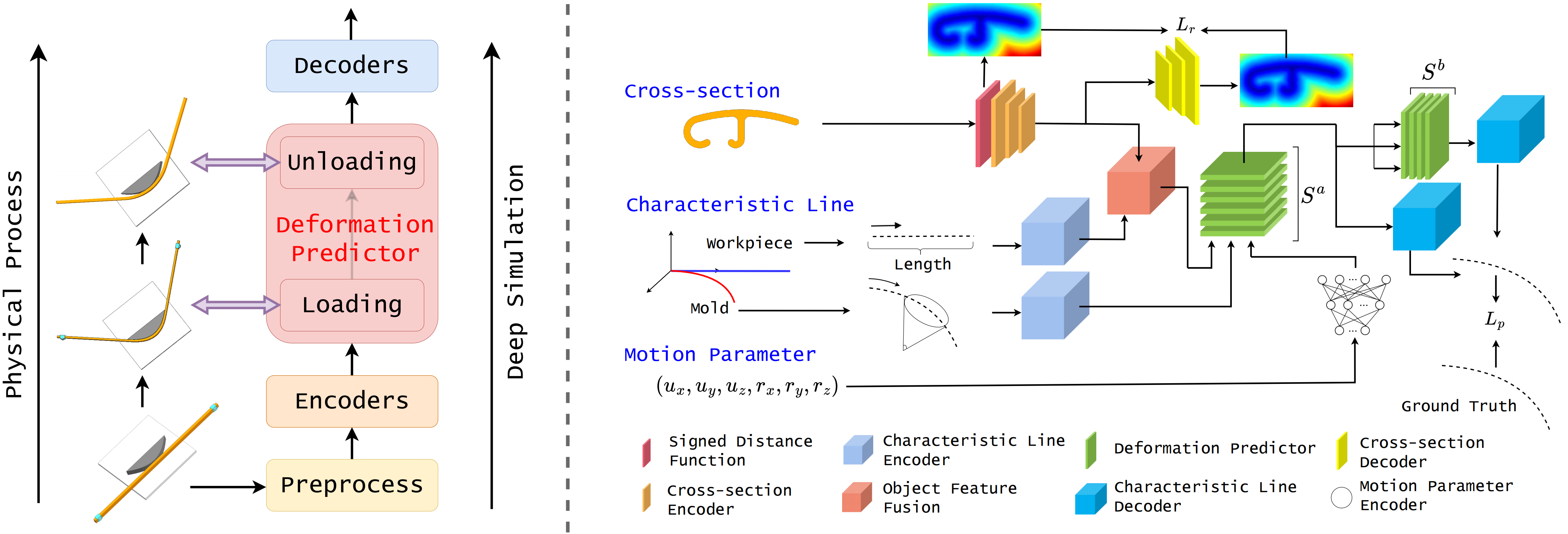}
  \caption{Overall design of LaDEEP. The left is the framework overview. The design of LaDEEP follows the physical process of stretch bending of metal, including two physical stages: loading and unloading. The right is the detailed model structure.}
  \label{figure3}
\end{figure*}

\begin{itemize}
    \item We propose LaDEEP, a novel deep learning framework tailored for an industrial instance, stretch bending task, with modules designed to consider its key order properties.
    \item We introduce the Deformation Predictor (DP), a two-stage Transformer-based module that effectively captures complex deformation behaviors through cross-attention and self-attention mechanisms.
    \item We generate a new dataset for our case, supporting our approach and filling a critical data gap in this domain. Experiments show that LaDEEP achieves five magnitudes faster speed than classical FEM method \cite{reddy2019introduction} with a comparable accuracy and gains 20.47\% relative improvement on average compared to deep learning baselines across all metrics.
    \item We complete the design of seven products in practical scenario. Two of them have already been put into real production with average mean absolute distance (MAD) $0.305mm$.
\end{itemize}

\section{Related Work}

Recently, numerous data-driven deep learning models have been widely employed to tackle various challenges in continuum mechanics, demonstrating remarkable potential in predicting material responses, simulating complex mechanical behaviors, and improving computational efficiency.

\textbf{Physics-Informed Neural Networks } These approaches formulate the PDEs, including governing equations, source items, initial conditions, and boundary conditions, as loss functions within deep models and many of them have been demonstrated on solid problems \cite{raissi2019physics, haghighat2021physics, li2023pre, he2023deep, lechner2024physically, xiong2024physics, yang2021deep}. During training, the output of the model progressively conforms to the PDE constraints, ultimately providing an accurate approximation of the PDE solution. However, these approaches require the precise formula of PDEs in solving forward problem, making them difficult to apply to real-world scenarios with incomplete observations. Additionally, they are typically limited to solving a single problem instance, and any change in parameters requires retraining the models.

\textbf{Neural Operators } The idea of neural operators is to learn mappings between two infinite-dimensional function spaces. The most prevailing model is FNO \cite{li2022fourier}, which approximates the integration with linear projection in the Fourier domain. Building on this, various variants \cite{wen2022u, rahman2022u, gupta2021multiwavelet, tran2021factorized, li2020neural, li2023fourier, li2024geometry, bonev2023spherical} have been proposed with crafted architectures to improve the accuracy, efficiency and application extensions. DeepONet \cite{lu2019deeponet} is also a prevalent model for operator learning which is designed based on the Universal Approximation Theorem for Operator \cite{chen1995universal}. However, as shown in our experiments, they tend to degenerate in our case, which involves non-smooth boundary conditions (solid contacts) and multiple stages (unloading and loading) due to the lack of consideration over the properties of solids and staged modeling.

\textbf{Transformer-based PDE Solvers } Transformers \cite{vaswani2017attention} have also been widely employed to solve PDEs \cite{liu2024mitigatingspectralbiasmultiscale, xiao2024improvedoperatorlearningorthogonal, li2024harnessing, zhou2024unisolver, li2023transformerpartialdifferentialequations, hao2023gnotgeneralneuraloperator}. Particularly, HT-Net \cite{ma2022ht} combines Swin Transformer \cite{liu2021swin} with multigrid methods \cite{wesseling1995introduction} to capture multiscale spatial relationships. FactFormer \cite{li2024scalable} enhances efficiency by leveraging a low-rank structure with multidimensional factorized attention. LSM \cite{wu2023solving} is introduced to address the high-dimensional PDEs by utilizing spectral methods \cite{gottlieb1977numerical} within latent space. Transolver \cite{wu2024transolver} introduces physics-attention to dynamically partition the discretized domain into learnable slices that capture the underlying physical states. However, these general approaches struggle to handle our case due to the lack of the inherent property of the large deformations of elastic-plastic solids.

\section{Method}

The overview of LaDEEP is shown in Figure~\ref{figure3}. For a given stretch bending problem, the inputs contain a 3D-shaped workpiece, a 3D-shaped mold, and the motion parameters of the working arms. The output is the final shape of workpiece. Initially, we preprocess the data to structure the inputs and output, and design several encoders to extract property-aware tokens. We propose a two-stage, Transformer-based \cite{vaswani2017attention} module, the Deformation Predictor (DP), to effectively characterize physical interactions between objects. Finally, we decode high-dimensional tokens back to the original space to calculate losses and optimize different parts of the network.

\subsection{Preprocess}
The formats of the considered objects in our case are unstructured and unsuitable for deep learning models. We first need to preprocess these objects to organize their structure, reduce redundancy, and ensure completeness. In practice, most workpieces are slender with a constant cross-sectional shape along their length. This means the cross-section remains the same when cut parallel to certain faces of the workpiece, allowing it to be recorded as a 2D image. We define curves along the length of the workpiece as ``characteristic lines'', which contain curvature information and describe how the cross-section expands. A 2D cross-section and a characteristic line can fully represent the 3D workpiece. For the 3D mold, its cross-section is designed based on the workpiece, so it can be represented simply by its characteristic line. We sample each characteristic line (for both the workpiece and the mold) as a point set $\mathbf{p}=\{p_i|i=1,\cdots,M\}$, consisting of $M$ points where each point $p_i$ is a position vector of $(x, y, z)$. Regarding the motion of each working arm, it applies loads on the workpiece through moving and rotating, controlled by 6 degrees of freedom (DoFs) $\{u_{x}, u_{y}, u_{z}, r_{x}, r_{y}, r_{z}\}$, where $\{u_{x}, u_{y}, u_{z}\}$ represent spatial displacements, and $\{ r_{x}, r_{y}, r_{z}\}$ are rotations around each axis. In practice, the middle of workpiece is fixed when processing, and two working arms move independently to process each half. Therefore, without violating the physics, we can independently focus on the deformation of one side and combine two sides completely \cite{lechner2024physically}.

\subsection{Encoder}

After processing the objects, we introduce several distinct encoders to individually encode each object. These encoders are designed based on the properties of different objects and integrate those properties into the token sequences.

\begin{figure}[h]
  \centering   
\includegraphics[width=0.8\linewidth,keepaspectratio]{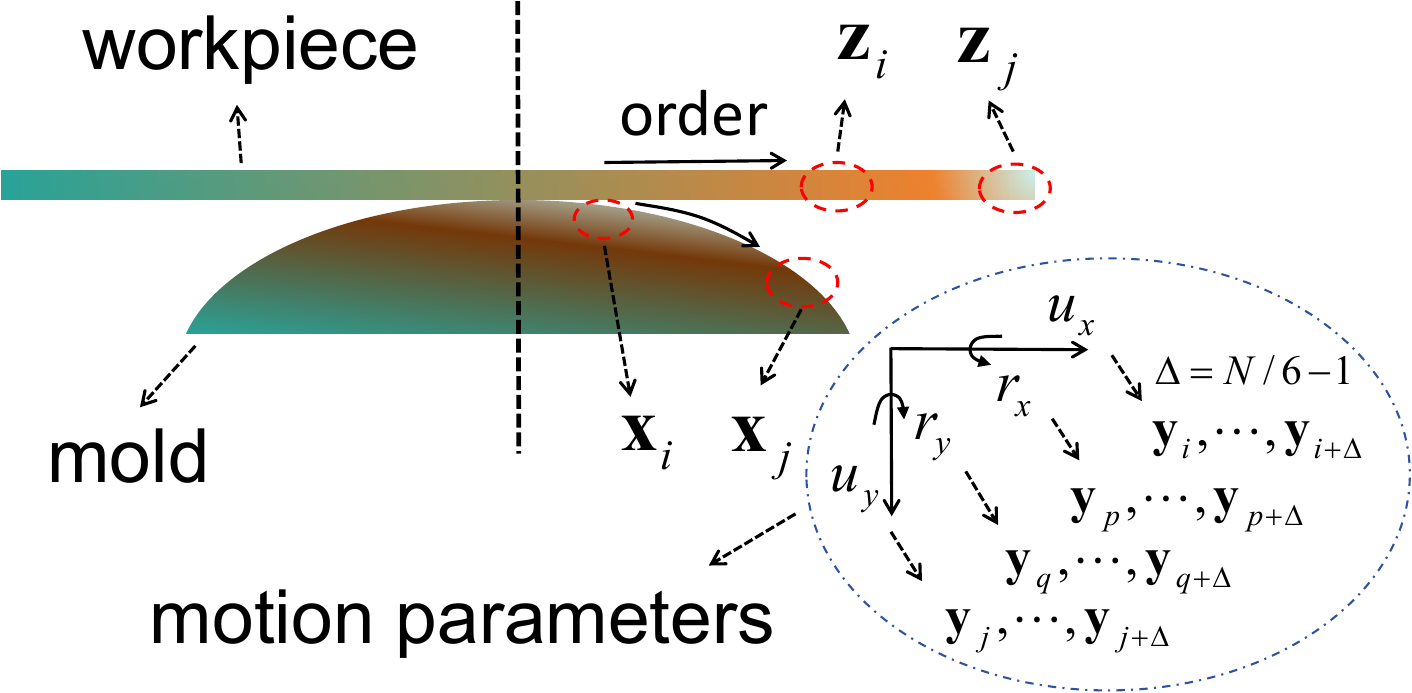}
     \caption{Property-aware tokens denoted as sequence of $\mathbf{x}$ and $\mathbf{z}$ corresponding to the regions of mold and workpiece.}
 \label{tokens}
 \vspace{-5pt}
\end{figure}

\textbf{Characteristic Line Encoder (CLE)}. The characteristic line is sampled as a point set. As mentioned earlier, most workpieces are slender, and the molds follow a similar structure. Hence, the point set sampled from the characteristic line possesses an inherent property of order, which is different from point clouds and crucial for capturing different deformation behaviors. As shown in Figure~\ref{tokens}, we encode the line into sequential region tokens, ensuring each token ($\mathbf{x}$ and $\mathbf{z}$) has explicit order implications during modeling. First, we embed each point into a high-dimensional space with an embedding size of $C$ using a linear layer. We then patchify these high-dimensional points into $N$ regions, each with $M / N$ adjacent points ($M$ is set as an integer multiple of $N$). Next, $N$ distinct linear layers are utilized to separately project these local regions into tokens $\mathbf{x}_l\in\mathbb{R}^{N\times C}$. However, these tokens capture local curvature but lack global features, such as the overall trend and length. To incorporate global features, we utilize two linear layers to embed and encode $M$ original points into a global feature $\mathbf{x}_g \in \mathbb{R}^{1\times C}$. Then this global feature is repeated $N$ times and added to $\mathbf{x}_l$, forming the final token sequence. We apply two separate CLEs, with non-shared parameters, to encode the characteristic lines of both the workpiece and the mold, resulting in outputs ${\mathbf{x}_{w}}, {\mathbf{x}_{m}}\in\mathbb{R}^{N\times C}$, respectively.

\textbf{Cross-Section Encoder (CSE)}. Each workpiece has a constant cross-section that is described as a 2D image with shape $1\times H\times W$. We use the Signed Distance Function (SDF) \cite{guo2016convolutional} to provide a low-redundancy representation of the shape and structure for cross-section, omitting unnecessary brightness, spectrum and semantic information. Specifically, a contour set $Con$ in a domain $\Omega \subset \mathbb{R}^2$ is defined as $Con=\{(i,j)\in \mathbb{R}^2:g(i,j)=0\}$, where $g$ is the sign function indicating the point's position relative to the contour: $g(i,j)=-1$ when $(i,j)\in \Omega^\circ$ and $g(i,j)=1$ when $g(i,j)\in \complement_{\mathbb{R}^2}\Omega$. The SDF $D(i, j)$ is formulated as $D(i,j)=\min_{(i',j')\in Con}|(i,j)-(i'-j')|\cdot g(i,j)$, which measures the distance of a given point $(i,j)$ to the nearest contour point with a sign indicting the relative position. After getting the SDF representation $\mathbf{s}\in\mathbb{R}^{1\times H\times W}$, a frozen pre-trained ResNet \cite{he2016deep} is utilized as a backbone to extract features, commonly applied in Computer Vision (CV) \cite{voulodimos2018deep}. Since the backbone is trained on natural images and may not be suitable for the SDF, we add convolutional layers with trainable parameters for greater flexibility. The cross-section of the workpiece is taken as input, resulting in the flattened output feature $\mathbf{s}_w\in\mathbb{R}^{1\times C}$.

\textbf{Object Feature Fusioner (OFF)}. Recall that a workpiece is described by two separated features, the cross-section $\mathbf{s}_w\in\mathbb{R}^{1\times C}$ and the characteristic line $\mathbf{x}_w\in\mathbb{R}^{N\times C}$. We then fuse them to obtain complete workpiece representation. We repeat the feature $\mathbf{s}_w$ by $N$ times and add it into $\mathbf{x}_w$ to form the fused result $\mathbf{z}_w\in\mathbb{R}^{N\times C}$. In this way, on the one hand, we can ensure that each token in $\mathbf{x}_w$ contains cross-sectional information, preserving the knowledge of constant cross-section. On the other hand, it does not disrupt the inherent order property within the feature of characteristic line $\mathbf{x}_w$.

\textbf{Motion Parameter Encoder (MPE)}. As claimed before, the loads applied on the workpiece are caused by the movement of each working arm, which is controlled by 6 DoFs. We use multiple tokens to capture the comprehensive influence of the movement effect along each degree of freedom. Specifically, the DoF of each axis is embedded with size $C$ and then separately projected into $N = Y \times 6$ tokens in total by 6 distinct linear layers ($N$ is set as an integer multiple of $6$). Each DoF is represented by adjacent $Y$ tokens, distributing the effects across different tokens for better learning. The final output are the tokens $\mathbf{y}_m\in\mathbb{R}^{N\times C}$, representing the motion parameters.

We finally obtain three property-aware sequences corresponding to the mold $\mathbf{x}_m$, the motion parameters $\mathbf{y}_w$ and the workpiece $\mathbf{z}_w$. They share the same shape with  $N\times C$, where $N$ is the number of tokens and $C$ is the embedding size. Without introducing ambiguity and for simplicity, we remove the subscript notation and use $\mathbf{x}$, $\mathbf{y}$, $\mathbf{z}$ to represent these 3 sequences, respectively. Figure~\ref{tokens} illustrates the stretch-bending process on the $xoy$ plane and indicates the corresponding tokens. Each object’s modeling retains its distinct physical structure, and each token contains specific inherent properties.

\begin{figure*}[t]
    \centering
    \includegraphics[width=\linewidth,keepaspectratio]{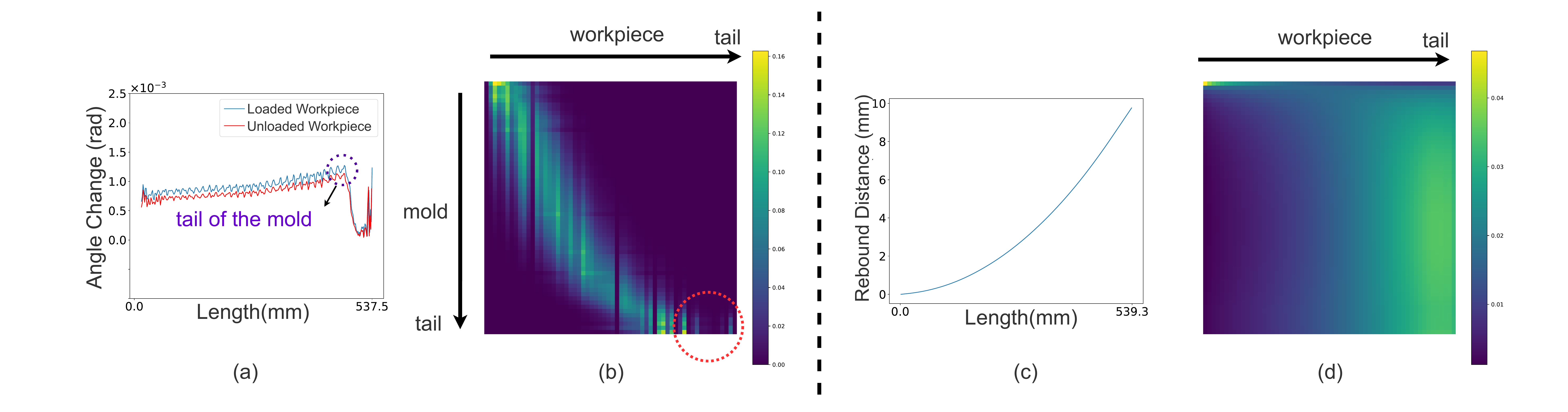}
    \caption{Visualization of attention maps. (a) The angle change along the length of the workpiece. (b) The contact points will be assigned higher weights. The red circle refers to the part of the workpiece that exceeds the mold. (c) The rebound distance along the length of the workpiece. (d) The weights change smoothly. Higher weights are assigned to parts with more rebound.}
    \label{attention_maps}
\end{figure*}

\subsection{Deformation Predictor}
\label{Deformation Predictor}

In traditional FEM method, the process is treated locally with incremental approximation. However, with the powerful nonlinear modeling capability of deep models, we can approximate solutions globally, accounting for interaction across the entire structure. As we have encoded the sequentially partitioned objects into property-aware token sequences, which contain essential order property of corresponding solids, it is natural to leverage the Transormer \cite{vaswani2017attention}, which excels at capturing global relationships and long-range dependencies, to explore the global interactions with the order property. Hence, we propose the Deformation Predictor (DP) -- a two-stage, Transformer-based module, to  effectively capture complex interactions between objects and approximate solutions with property-aware token sequences.

In the first stage, we employ Transformer with cross-attention mechanism \cite{chen2021crossvit} to model the relationships between objects in the loading stage. During this process, the working arm moves and rotates, exerting loads on the workpiece, causing it to lengthen and curve over a mold. Both the motion parameters and the mold affect the deformation of the workpiece through movement along 6 DoFs and complex nonlinear, non-smooth contact.  Recall that $\mathbf{x}$, $\mathbf{y}$ and $\mathbf{z}$ respectively represent the property-aware token sequences of the motion parameters, the mold and the workpiece, the relationships are modeled in the $i$-th layer as:
\begin{equation}
\begin{aligned}
    (\mathbf{x}_0,\mathbf{y}_0,\mathbf{z}_0)&=(\mathbf{x}+\mathbf{x}_{pos},\mathbf{y}+\mathbf{y}_{pos},\mathbf{z}+\mathbf{z}_{pos})\\
    \mathbf{q}_{i-1}&=\text{Linear}(\text{concat}(\mathbf{x}_0,\mathbf{y}_0))\\
    \mathbf{k}_{i-1},\mathbf{v}_{i-1}&=\text{Linear}(\mathbf{z}_{i-1}),\text{Linear}(\mathbf{z}_{i-1})\\\mathbf{z}'_{i-1}&=\text{LN}(\mathbf{z}_{i-1}+\text{softmax}(\frac{\mathbf{q}_{i-1}\mathbf{k}_{i-1}^T}{\sqrt{C}})\mathbf{v}_{i-1}) \\
\mathbf{z}_{i}&=\text{LN}(\mathbf{z}'_{i-1}+\text{FFN}(\mathbf{z}'_{i-1}))
\end{aligned}
\end{equation}
Here, $\mathbf{x}_{pos},\mathbf{y}_{pos},\mathbf{z}_{pos}\in\mathbb{R}^{N\times C}$ are position embeddings. Linear($\cdot$) is a linear projection layer, and LN($\cdot$) is a normalization layer. The $j$-th row of $\mathbf{q}$ contains information of the $j$-th region in the mold and the $\lfloor j / Y \rfloor$-th DoF of the motion parameters. The $j$-th rows of $\mathbf{k}$ and $\mathbf{v}$ represent the $j$-th region in the workpiece. Consequently, through the cross-attention, the element of the $j$-th row and the $k$-th column of the attention weight matrix represents the influence of the combined action of the $\lfloor j / Y \rfloor$-th DoF's motion and the $j$-th region of the mold on the $k$-th region of the workpiece. And thus, \emph{the attention mechanism properly characterizes physical interaction between the mold and workpiece, which rationalizes the usage of Transformer architecture}.  There are $S^a$ layers in total. Throughout the process, both the mold and the motion parameters remain invariant. Therefore, we use the concat($\cdot$) of the initial $\mathbf{x}_0$ and $\mathbf{y}_0$ as inputs for each layer, only forwarding $\mathbf{z}_0$. The final output, $\mathbf{z}^a\in\mathbb{R}^{N\times C}$, is the high-dimensional representation of the workpiece after loading.

In the second stage, we use Transformer with self-attention mechanism \cite{vaswani2017attention} to learn the rebound of the workpiece. During this process, the working arm is released, allowing the elastic part of the deformation in the workpiece to recover, causing a rebound. Since only the workpiece itself is involved, we utilize self-attention to model this stage as follows:
\begin{equation}
\begin{aligned}
    \mathbf{z}_0&=\mathbf{z}^a+\mathbf{z}^a_{pos}\\
    (\mathbf{q}_{i-1},\mathbf{k}_{i-1},\mathbf{v}_{i-1})&=\text{Linear}(\mathbf{z}_{i-1})\\ \mathbf{z}'_{i-1}&=\text{LN}(\mathbf{z}_{i-1}+\text{softmax}(\frac{\mathbf{q}_{i-1}\mathbf{k}_{i-1}^T}{\sqrt{C}})\mathbf{v}_{i-1})\\ 
\mathbf{z}_{i}&=\text{LN}(\mathbf{z}'_{i-1}+\text{FFN}(\mathbf{z}'_{i-1}))
\end{aligned}
\end{equation}
This is the forward in the $i$-th layer. We project $\mathbf{z}_0$ into $\mathbf{q},\mathbf{k},\mathbf{v}\in\mathbb{R}^{N\times C}$ as the inputs of self-attention. The $j$-th rows of $\mathbf{q}$, $\mathbf{k}$ and $\mathbf{v}$ contain information of the $j$-th region in the workpiece. Through self-attention, the element of the $j$-th row and the $k$-th column of the attention weight matrix represents the influence of the $j$-th region on the $k$-th region of the workpiece in the rebound stage. There are $S^b$ layers in total, and the final output, $\mathbf{z}^b\in\mathbb{R}^{N\times C}$, is the high-dimensional representation of the workpiece after unloading.

Toward an intuitive understanding of DP, we visualize the attention maps. There exists a noticeable corner (Figure~\ref{attention_maps}a) corresponding to the tail end of the mold, where the workpiece extends beyond the mold, resulting in a distinct abrupt change. As shown in Figure~\ref{attention_maps}b, the loading module learns the interaction pattern of the mold on the workpiece due to the explicit inherent order property in the input token sequences: the interaction is most evident at the contact points between the mold and the workpiece, which are assigned higher weights, while the part of the workpiece extending beyond the mold is minimally affected. Figure~\ref{attention_maps}c and \ref{attention_maps}d show that the rebound increases as the position becomes closer to the tail of the workpiece. This is consistent with the pattern learned by the unloading module: the attention weights change smoothly, with greater weights assigned as it is closer to the tail end.

\subsection{Decoder and Loss Function}
\label{Decoder and Loss Function}
After getting the high-dimensional representation of the workpiece, we develop several decoders to map the representation back to the original space and calculate losses.

\textbf{Characteristic Line Decoder (CLD)}. We roughly reverse the structure of the Characteristic Line Encoder (CLE) as Characteristic Line Decoder (CLD) to decode the high-dimensional representation back to the original space. Take $\mathbf{z}^b\in\mathbb{R}^{N\times C}$ from unloading module as input, we first use $N$ distinct linear layers to decode each token to adjacent $M / N$ points. Then, a linear layer is utilized to de-embed these points back to the Euclidean space. We decode $\mathbf{z}^a\in\mathbb{R}^{N\times C}$ from loading module in the same way. The final outputs for $\mathbf{z}^a$ and $\mathbf{z}^b$ are $\mathbf{p}^a\in\mathbb{R}^{M\times 3}$ and  $\mathbf{p}^b\in\mathbb{R}^{M\times 3}$, respectively. 

\textbf{Cross-Section Decoder (CSD)}. Given that the cross-section has a significant impact on the forming of the workpiece \cite{yu2018research}, we establish a Cross-Section Decoder (CSD) to recover SDFs, ensuring the extracted features by Cross-Section Encoder (CSE) are effective. We refer to the structure of VAE \cite{kingma2022autoencoding} and utilize deconvolution layers and interpolation operations to construct CSD. Take $\mathbf{s}^w\in\mathbb{R}^{1\times C}$ from CSE as input, the output is $\mathbf{s}_r\in\mathbb{R}^{1\times H\times W}$, matching the shape of ground truth $\mathbf{s}\in\mathbb{R}^{1\times H\times W}$.

\textbf{Loss Functions} Our approach utilizes two distinct loss functions: the reconstruction loss $loss_r$, for the cross-section, and the prediction loss $loss_p$ , for the workpiece. The $loss_p$ is calculated twice, once for the workpiece after loading and once after unloading. We use 3 optimizers to simultaneously train different parts of the network with corresponding losses. For the reconstruction loss, we measure the Mean Square Error (MSE) $loss_r=\frac{1}{n}\sum_1^n(\mathbf{s}-\mathbf{s}_r)^2$. The function $loss_p$ measures the discrepancy between the characteristic lines. Due to imbalanced value distributions across different coordinate axes of the 3D characteristic line, normalizing all axes equally could lead to incorrect shifts. To address this, we emphasize axes with more significant orders of magnitude by employing a coordinated L2 loss defined as: $loss_p=\lambda_x \|\Delta{x}\|_2+\lambda_y \|\Delta{y}\|_2+\lambda_z \|\Delta{z}\|_2$
, where $\lambda_x$, $\lambda_y$ and $\lambda_z$ are weights corresponding to 3D axes, computed based on the data range along each axis. The terms $\Delta{x}$, $\Delta{y}$ and $\Delta{z}$ denote the differences between the ground truth $\mathbf{\Tilde{p}}^a$ and $\mathbf{\Tilde{p}}^b$ and the prediction $\mathbf{p}^a$ and  $\mathbf{p}^b$ along each respective axis.

\section{Experiments}

We conduct extensive experiments to evaluate the effectiveness of Ladeep, including the comparison with numerical solvers and deep learning models, ablations, out-of-distribution generalization and real deployment.

\subsection{Experiment Settings}

\begin{figure}[h]
  \centering   
  \includegraphics[width=0.6\linewidth,keepaspectratio]{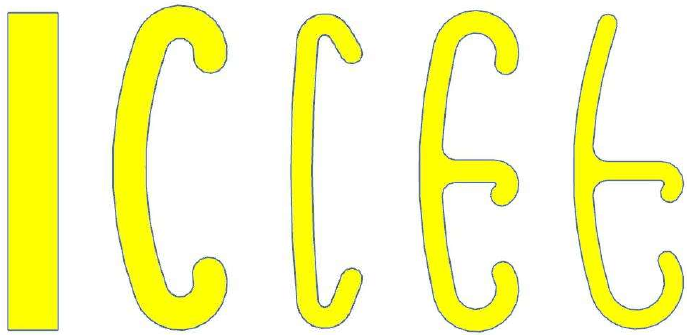}
    \caption{Five types of cross-section, which are indexed 1-5 from left to right.}
\label{figure4}
\vspace{-3pt}
\end{figure}

\textbf{Dataset}. We employ FEM \cite{reddy2019introduction} with fine mesh to generate highly accurate dataset. This dataset contains 3000 samples and each sample contains: a characteristic and a cross-section of the 3D workpiece, a characteristic of the 3D mold, and the motion parameters. For the cross-section, we select 5 representative types of cross-section structures, depicted in Figure~\ref{figure4}, which can cover most of the practical products. There are random parameters that control the arc radius, arc radian, thickness and height. For the characteristic line of the workpiece, we place the initial straight workpiece on the x-axis from original point to the maximal length and randomly sample the length. Regarding the mold, we utilize two 1/4 elliptical arcs on two perpendicular planes to combine a 3D curve as the characteristic line. The motion parameters are calculated by classical involute approach \cite{arnold2012classification} with the characteristic line of the mold. We use Abaqus \cite{khennane2013introduction, abaqus2011abaqus} software, a robust and widely adopted industrial simulation software based on FEM, to perform the computation. The dataset is split into an 8:1:1 ratio for training, evaluation and test. More details about the dataset are in Appendix \ref{Dataset Generation}.

\textbf{Baselines.} The compared baselines are classified into two groups, the traditional numerical methods, and the deep learning methods. For traditional methods, we select FEM as the representative. We utilize Abaqus \cite{khennane2013introduction, abaqus2011abaqus} as a comparable baseline. Each assembly model is divided into around 20,000 elements. we use scale factors \{0.8, 0.6, 0.4, 0.2\} to reduce the number of meshes in the highest resolution assembly models. During the loading stage, we employ the explicit dynamic algorithm \cite{barbero2023finite} for forward calculations based on dynamic equations. For the unloading stage, we use the implicit static algorithm \cite{barbero2023finite}, iteratively solving the problem with Newton's method \cite{kelley2003solving}. Since the explicit dynamic algorithm in Abaqus does not support GPU computation, we leverage a 32-core CPU server for parallel processing using MPI \cite{walker1996mpi}. We also assess the impact of different core counts on computation speed and find that 32 cores nearly reach the maximum acceleration for our single assembly model. For existing DL methods, we comprehensively compare LaDEEP with 10 well-known models: DeepONet \cite{lu2019deeponet}, FNO \cite{li2022fourier}, GINO \cite{li2024geometry}, SFNO \cite{bonev2023spherical}, TFNO \cite{kossaifi2023multi}, UNO \cite{rahman2022u}, FactFormer \cite{li2024scalable}, LSM \cite{wu2023solving}, Transolver \cite{wu2024transolver} and TCN \cite{oord2016wavenet} which are well-known models of PDE solvers. We implement all the deep learning baselines based on their official code. All the baselines are trained and tested under the same training strategy and loss functions as LaDEEP.

\textbf{Metrics}. We utilize three different metrics to evaluate the performance of the model from multiple aspects. The definition of these metrics are listed below:

\begin{itemize}
    \item \textbf{MAD}. The mean absolute distance (MAD) measures the characteristic line distance between the prediction and ground truth of the workpiece.
    \item \textbf{IoU 3D}. The intersection over union (IoU) of the reconstructed workpiece compared to the ground truth. We utilize it to assess the real prediction accuracy at each point.
    \item \textbf{Tail Error (TE)}. For large deformation, the errors on the tail faces are more likely to accumulate. We therefore employ Tail Error (TE) for evaluation. It measures the mean absolute error between point in the tail face for the prediction and ground truth of the workpiece.
\end{itemize}

Additionally, we use relative error reduction to compute the promotion portion w.r.t. the second best model, which is formulated as $\frac{\text{The second best error} - \text{Our error}}{\text{The second best error}}\times 100\%$.

\subsection{Main Results}
\label{Main Results}
\begin{figure}[htbp]
    \centering
    \includegraphics[width=0.8\linewidth,keepaspectratio]{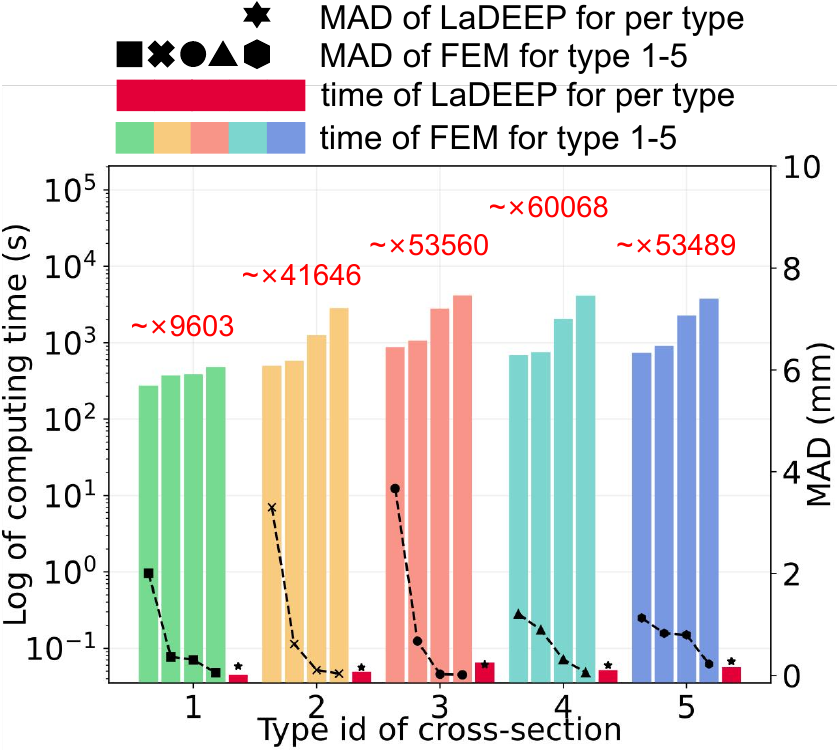}
\caption{Comparison between LaDEEP and FEM on the accuracy and speed, respectively.}
\label{figure1:b}
\end{figure}

\textbf{Compared to FEM } Figure~\ref{figure1:b} shows the results of computing time and MAD between LaDEEP and traditional FEM. The FEM methods are set with various granularities of meshes (coarser than that for dataset generation) for different computation efficiency and accuracy. For FEM, a reduced number of elements leads to improved computational speed, but at the expense of accuracy. For workpieces of the type-1 cross-section, FEM methods need around 342s for coarse-grained meshes, with errors on $2.0092mm$(MAD) and around 598s for finer-grained meshes, with $0.0568mm$(MAD). For LaDEEP, only $0.0453s$ is needed for computations with $0.1823mm$(MAD), providing around 9603 times acceleration. With the similar computation accuracy, the accelerations range from 9603 times to around 60068. In our dataset, the type-1 is the simplest, and type-4 is the most complicated one that has two arcs with different radius and arc length. The FEM needs much more computation time for type-4. But for LaDEEP, the shapes of cross-sections are all represented by SDFs with shape $512\times 256$. The inference speed is consistent regardless of shape complexity, as computation time does not increase for more intricate profiles. We therefore observe a higher acceleration ratio.

\textbf{Compared to deep learning alternatives } To evaluate LaDEEP from a comprehensive view, we experiment in two settings: (1) We compare LaDEEP with other naive baselines. These baselines are added with simple encoders and decoders constructed with linear layers. (2) We compare LaDEEP with other modified baselines. The encoders and decoders in LaDEEP are kept unchanged. We only replace the Deformation Predictor (DP) with these baselines. From these two settings, we can explore the effectiveness of the encoders, decoders and DP in LaDEEP.

\begin{table*}[htbp]
    \centering
    \caption{Performance comparison with baselines. Columns 2-4 are results in setting (1), and columns 5-7 are results in setting (2). For MAD and TE, a smaller value indicates better performance, whereas for IoU 3D, the opposite is true. The best result is in bold and the second best is underlined. Improvement in the last column refers to the average relative error reduction across all metrics of corresponding models. Improvement in the last row refers to the relative error reduction w.r.t the second best model.}
     \resizebox{0.85\textwidth}{!}{
    \begin{tabular}{c|ccc|ccc|c}
        \toprule
        & \multicolumn{3}{c|}{Setting (1)} & \multicolumn{3}{c|}{Setting (2)} \\
        \midrule
        Model &MAD(mm)& IoU 3D(\%)&TE(mm)&MAD(mm)& IoU 3D(\%)&TE(mm)&Improvement\\
        \midrule
        DeepONet & 0.3836 & 74.43 & 0.7806& 0.2445 & 82.52 & 0.6238&22.41\%\\
        FNO & \underline{0.3251} & 78.15 & 0.7488 & 0.2325 & 82.27 & 0.6060&16.44\%\\
        GINO & 0.3394 & 76.65 & 0.8067 & 0.2567 & 81.08 & 0.6812&15.24\%\\
        SFNO & 0.3366 & 77.79 & 0.7737 & 0.2362 & 82.26 & 0.6079&19.07\%\\
        TFNO & 0.3267 & 78.03 & 0.7598 & 0.2389 & 81.59 & 0.6419&15.65\%\\
        UNO & 0.3380 & 77.56 & 0.7795 & 0.2137 & 83.33 & 0.5810&23.23\%\\
        FactFormer & 0.3458 & \underline{78.28} & 0.8134 & 0.2404 & 82.17 & 0.6478&18.60\%\\
        LSM & 0.3356 & 77.86 & \underline{0.7222} & 0.2099 & 84.00 & \underline{0.4987}&25.43\%\\
        Transolver & 0.3912 & 76.31 & 0.8359 & \underline{0.2052} & \underline{84.21} & 0.5422&31.03\%\\
        TCN & 0.4047 & 73.12 & 0.8080 & 0.2585 & 78.61 & 0.6431&21.35\%\\
        \midrule
        LaDEEP & \textbf{0.1698} & \textbf{86.58} & \textbf{0.4591} & \textbf{0.1698} & \textbf{86.58} & \textbf{0.4591} & /\\
        Improvement & 47.77\% & 10.64\% & 36.43\% & 17.25\% & 2.81\% & 7.94\% & /\\
        \bottomrule
    \end{tabular}}
    \label{table1}
\end{table*}

\label{Compared to DL models}
  \begin{figure}[htbp]
  \centering
            \includegraphics[width=\linewidth,keepaspectratio]{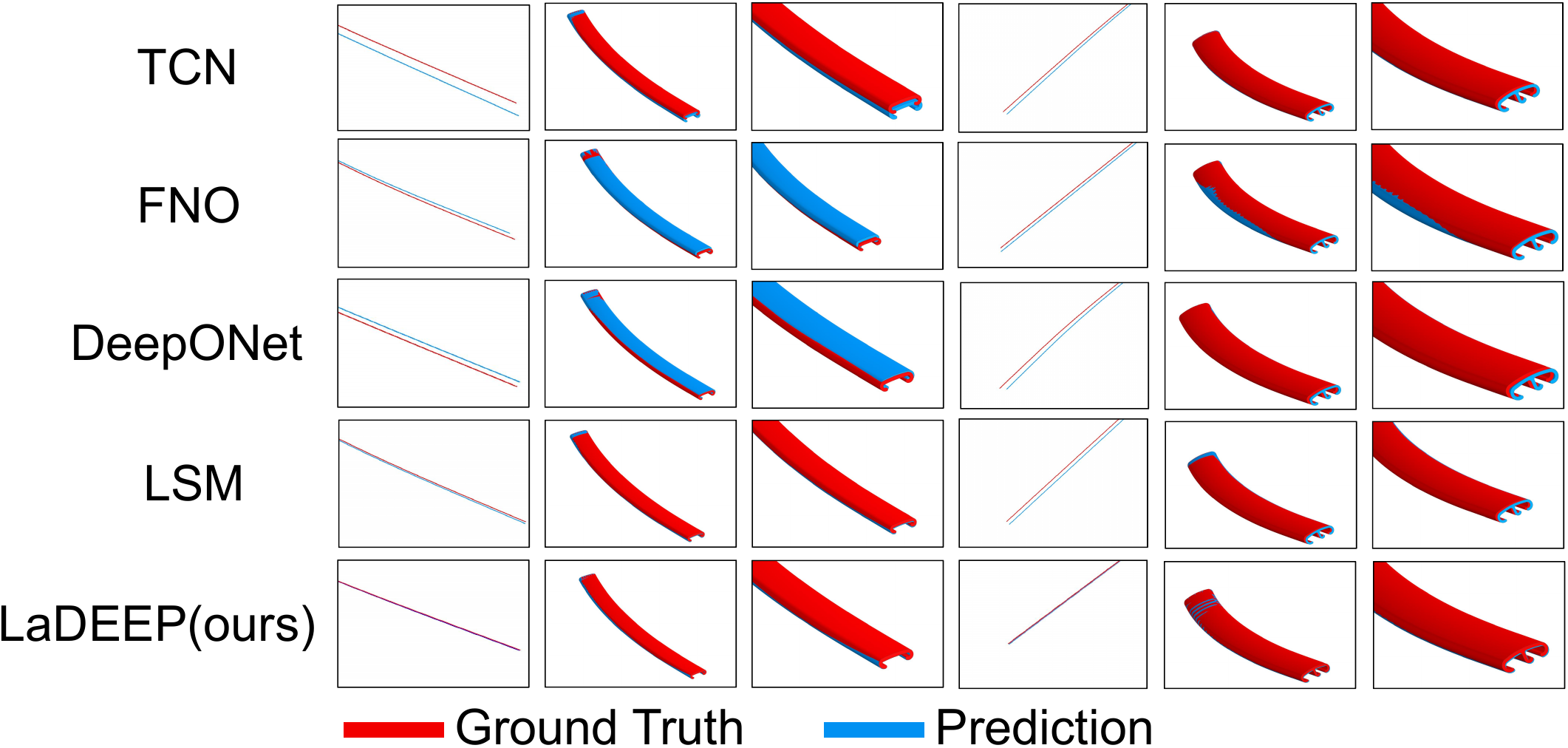}
		\caption{Visualization of results in setting (2). Column 1 and 4 are characteristic lines. Column 2-3 and column 5-6 are workpieces corresponding to column 1 and 4, respectively. We only select part of results to visualize due to the space limitation.}
		\label{figure5}
\end{figure}

As presented in Table~\ref{table1}, LaDEEP performs consistent state-of-the-art in both settings. Notably, from setting (1) to (2), these baselines also gain significant improvement (20.85\% on average) with proposed encoders and decoders, demonstrating the effectiveness of our design in handling large deformations of elastic-plastic solids. Also, some advanced Transformer-based models, such as LSM and Transolver, achieve impressive improvement (25.43\% and 31.03\%) after taking property-aware token sequences as inputs. This is because these models are designed for general cases, without considering problem-specific properties, such as the order of the sequentially partitioned regions on the slender workpiece. After incorporating the property-aware token sequences, these models are more effective to characterize the interactions between objects. Additionally, they perform better then neural operators due to the strong capacity of Transformer in global modeling and long-distance dependency. However, the considered case is involved two stage -- loading and unloading. Without staged modeling approach, they struggle to approximate the solutions accurately. Neural operators also face the same dilemma, although they also raise some improvements with our designed modules. As shown in Figure~\ref{figure5}, the deviations in the predictions are primarily due to discrepancies in length (row 2, column 1-3) and shifts in curve positions (row 2, column 4-6). Benefit from analyzing the inherent properties of the case, LaDEEP can effectively capture underlying correlation between objects. We thereby achieve superior performance among all models.

\textbf{Out-of-distribution (OOD) Generalization } To further explore the model's transfer ability, we conduct further experiment to assess its potential in transferring to other tasks. We first conduct experiments to explore the generalization of LaDEEP to unseen cross-sections. We keep the training parameters the same for all experiments except the training epoch. The training setting are: (1) \textbf{Zero-shot Testing}: We use the first 4 out of 5 kinds of cross-section as training data and the 5th unseen cross-section as test data. The model is trained for 600 epochs.(2) \textbf{Fine-tuning}: We further investigate by splitting the 5th cross-section into an 8:2 ratio for training and test. We fine-tune the model in (1) for additional 200 epochs. (3) \textbf{Full Dataset Training}: We evaluate the model on the 5th cross-section using the model trained with all types of cross-sections. The results are shown in Table~\ref{table_unseen_sections} which demonstrate that while the LaDEEP exhibits some zero-shot capabilities, there is room for improvement. Fine-tuning significantly enhances performance, achieving results close to the baseline model trained with all cross-sections. This highlights LaDEEP's potential in practical applications. When data with unseen cross-section appears, fine-tuning the model to generalize the data is a considerable way. 

\begin{table}[h]
    \centering
    \caption{Results of Cross-section Generalization.}
    \resizebox{0.8\linewidth}{!}{
    \begin{tabular}{c|ccc}
        \toprule
        Setting ID & MAD(mm) &IoU 3D(\%) & TE(mm)\\
        \midrule
        1 & 0.4122 & 72.02 & 1.1084\\
        2 & 0.2232 & 84.56 & 0.6153\\  
        3& 0.2313 & 83.78 & 0.6341\\
        \bottomrule
    \end{tabular}}
    \label{table_unseen_sections}
\end{table}

Except for the cross-section, we also explore the model's generalization across different metal materials. We use a new aluminum alloy with different alloy ratios compared to the original dataset. These alloys differ in material parameters: hardness, density, Poisson's ratio, Young's modulus, and stress-strain behavior, leading to significant differences in deformation, stress, and rebound behavior. The parameters for both alloys are measured from materials used in practical production. Using the 5 cross-sections mentioned in the paper, we generate a total of 300 data samples, which are split into an 8:2 ratio for training and test. Two experiment settings: (1) We take the model in the paper as the basic pre-trained model with is trained on the original dataset for 600 epochs. Then we test the new data directly on the pre-trained model. (2) We fine-tune the pre-trained model in (1) for 200 epochs with the new dataset. The results are presented in Table~\ref{table_new_materials}.

\begin{table}[h]
    \centering
    \caption{Results of Cross-section Generalization.}
    \begin{tabular}{c|ccc}
        \toprule
        Setting ID & MAD(mm) &IoU 3D(\%) & TE(mm)\\
        \midrule
        1 & 0.2053 & 84.43 & 0.5338\\
        2 & 0.1711 & 85.56 & 0.4389\\  
        \bottomrule
    \end{tabular}
    \label{table_new_materials}
\end{table}

These results demonstrate that with our pre-trained model, we can achieve rapid convergence on new alloy data using only a small dataset and minimal fine-tuning iterations. This capability is highly valuable in practical applications where adapting to new materials efficiently is crucial.

\subsection{Ablation Studies} \label{Ablation Studies}

The proposed model comprises several essential components and we assess their efficacy through comprehensive ablation studies. We consider four distinct variants including encoders, fusioner, predictor and loss function. The settings are: 1) Replace the Object Feature Fusion (OFF) with attention mechanism which is also a global fusioner; 2) Replace the SDF with gray image. 3) Replace the CLE with PointNet \cite{qi2017pointnet} containing max operation that omits the order information 4) Replace the 1st stage of the DP with MLP \cite{pinkus1999approximation} to determine whether the global modeling and long-distance dependency are important; 5) Replace the 2nd stage of the DP with MLP; 6) Replace the $loss_p$ with Mean Square Error (MSE) which gives equal equations to all axes.

\begin{figure*}[ht]
  \centering
  \includegraphics[width=\linewidth,keepaspectratio]{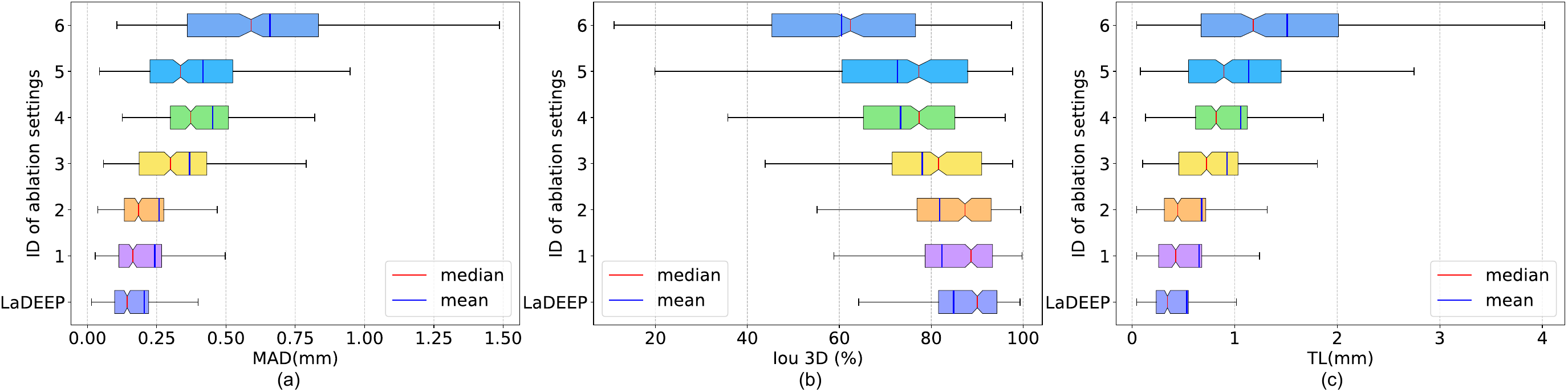}
  \caption{Results of ablation studies. The indexes on the y-axis are corresponding to the ablation setting indexes. For MAD and TE, a smaller value indicates better performance, whereas for IoU 3D, the opposite is true.}
  \label{figure7}
\end{figure*}

As shown in Figure~\ref{figure7}, our proposed essential modules synergistically boost the modeling capacity and performance. Compared with attention-based fusioner, while the accuracy is close, the computation times are $67.9ms$ (LaDEEP) and $86.1ms$ (attention-based) with batch size 8, respectively. The extra $26.8\%$ runtime penalty does not pay off. Using gray images of cross-sections as input lowers down the accuracy by $25.77\%$ on MAD caused by the redundant information of gray images. The consideration about the order of the points sampled from the characteristic line is significant. The expanded overall error distribution in the results indicates a shift in the implicit physics learned by the model. We can observe the same phenomenon if we replace two modules in DP with MLP, respectively. Due to the lack of the explicit physical modeling, the inductive bias of the model has shifted. This shift can lead to the model capturing incorrect dependencies and not learning accurate physical knowledge. The results of MSE loss are much worse. Note that the characteristic line is 3D. The coordinate components on 3 axes are imbalanced. The results of MSE loss proves that the coordinated L2 loss can effectively alleviate this problem.

\subsection{Deployment}
\label{Deployment}

\begin{figure}[htbp]
  \centering           
  \includegraphics[width=\linewidth,keepaspectratio]{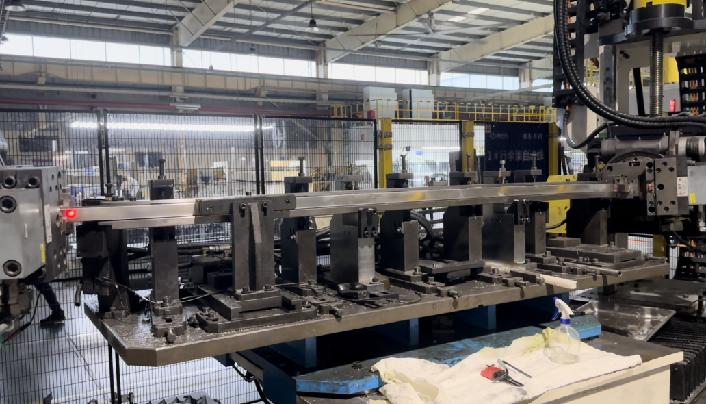}
		\caption{A mold designed based on LaDEEP is used for on-site production.}
  \label{figure8}
\end{figure}

We deploy LaDEEP into a real manufacturing factory with aluminum stretch bending fabrications. The FEM method has been used for computations over years but provides limited value due to the bad computation efficiency -- a practical workpiece requires days for a single FEM computation and the whole design process (iterative simulation) holds for 2-3 months. This gap motives us to design more efficient, deep learning-based surrogate models. 

In real deployments, there are always inevitable errors between the computations and productions due to the inherent complexities and uncertainties in the manufacturing process. These errors vary from different product scenarios with many sources such as the mold production error, the mold installation error, the motion parameters error, the material property variation error and etc. These errors can be additive or canceling, resulting an overall application error (AE) around $1mm\sim 10mm$ (e.g., a workpiece with length 2m, the relative error is around 0.5\%). Then, such application error will be compensated by empirical production techniques and feedback to computations, ensuring the final production error (PE) within 1$mm$. The empirical production techniques include machining of the mold by tools directly, and fine-tuning working arms. 

We deploy LaDEEP into practical scenarios and develop a two-cycled mold design paradigm as shown in Figure~\ref{two-loop} based on LaDEEP. More details are described in Appendix \ref{Two-Loop Mold Design Paradigm}. We accomplish seven real-product designs with LaDEEP and the average application error is around 8.5mm. This is sufficient for on-site adjustment. Two of them are corrected by on-site adjustment and the final production errors are shown in Table~\ref{table2}. The mold design process is reduced to around 1 week and the efficiency is improved by around 8-10 times. Figure~\ref{figure8} is a mold designed based on LaDEEP used for on-site production.

\begin{table}[htbp]
    \centering
    \caption{PE of two products designed by our two-loop mold design paradigm.}
    \resizebox{0.7\linewidth}{!}{
    \begin{tabular}{c|cc}
    \hline
    Product&DX11-RQT&H93-FUQT \\
    \hline
    MAD(mm)&0.32&0.29 \\
    \hline
    \end{tabular}}
    \label{table2}
\end{table}

\section{Conclusion} \label{Conclusion and Discussion}
In this paper, we propose LaDEEP, a novel deep learning-based framework tailored for an industrial task, stretch bending. We design several modules to encode sequential property-aware tokens and propose a two-stage, Transformer-based module, the Deformation Predictor (DP) to approximate the two-stage solutions. We generate a dataset to support our approach and fill the data gap in this area. LaDEEP achieves five magnitudes faster speed than FEM with a comparable accuracy, and gains 20.47\% relative improvement on average compared to other deep learning baselines. 

\section{Acknowledgments}
This work is supported partly by National Key Reasearch Plan under grant No.2021YFB2900100, the National Natural Science Foundation of China (NSFC) 61925202, and the Jiangsu Provincial Key Research and Development Program under Grant BE2022065-1, BE2022065-3.

\newpage

\bibliographystyle{ACM-Reference-Format}
\balance
\bibliography{sample-base}

\appendix

\section{Dataset Generation}
\label{Dataset Generation}
The training of deep learning models requires sufficient amount of data, which can hardly be obtained from real manufacturing environment. We employ traditional FEM method with fine mesh resolution to generate highly accurate dataset. Each sample is composed of three components, the 3D workpiece, the 3D mold, and the motion parameters. We collect both shapes of the workpieces after loading and unloading.

\begin{figure*}[htbp]
\begin{minipage}[t]{0.45\linewidth}
    \centering
    \includegraphics[width=\linewidth]{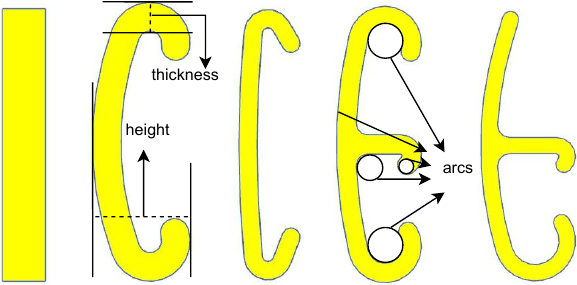}
    \caption{Five representative types of cross-section from practice. They are indexed 1-5 from left to right. Each kind of cross-section has different number, radian and radius of arcs, various thicknesses and heights.}
    \label{data_cross_section_appendix}
\end{minipage}
\hspace{0.08\linewidth}
    \begin{minipage}[t]{0.45\linewidth}
        \centering
    \includegraphics[width=\linewidth]{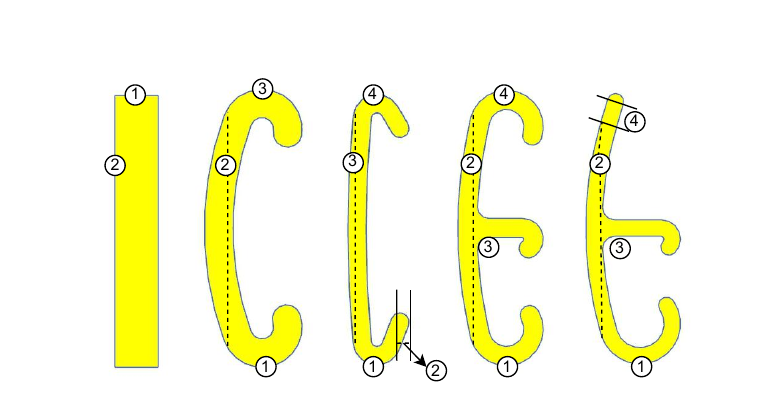}
    \caption{The indexes on each kind of cross-section are marks indicating that some parameters of the corresponding parts are sampled from specific distributions. The specific distributions are listd in Table~\ref{random_distributions}.}
    \label{data_cross_section_distribution}
    \end{minipage}
\end{figure*}

\begin{table*}[htbp]
    \centering
    \caption{Distributions that control the parameters for each kind of cross-section. The indexes in the table are corresponding to those in Figure~\ref{data_cross_section_distribution}. The whole cross-section can be calculated through given parameters. Let the positive direction of the x-axis be $0^\circ$ and the clockwise is the positive direction. The postfix "s" means start arc and "e" means end arc.}
    \begin{tabular}{c|ccc}
    \toprule
        Type ID & Radius Distribution (mm) & Radian Distribution ($^\circ$) & Length Distribution (mm)
    \\\midrule
        1 & / & / & \makecell[c]{thickness: $\mathcal{U}[2.5, 4]$\\1: $\mathcal{U}[2.5, 4]$; 2: $\mathcal{U}[17, 20]$}\\
    \midrule
    2 &  \makecell[c]{1: $\mathcal{U}[2.5, 3]$} & \makecell[c]{1s: $\mathcal{U}[95, 110]$; 1e: $\mathcal{U}[280, 310]$\\3s: $\mathcal{U}[240, 270]$} &\makecell[c]{thickness: $\mathcal{U}[1.8, 2.2]$\\2: $\mathcal{U}[14, 16]$}\\
    \midrule
    3 &  \makecell[c]{1: $\mathcal{U}[1.4,1.6]$} & \makecell[c]{1s: $\mathcal{U}[95, 110]$; 1e: $\mathcal{U}[230, 250]$\\4s: $\mathcal{U}[290, 310]$} &\makecell[c]{thickness: $\mathcal{U}[0.8, 1.2]$\\2: $\mathcal{U}[0.4, 0.6]$; 3: $\mathcal{U}[14, 16]$}\\
    \midrule
    4 &  \makecell[c]{1: $\mathcal{U}[2.5,3]$\\3: $\mathcal{U}[1.5, 1.7]$} & \makecell[c]{1s: $\mathcal{U}[95, 110]$; 1e: $\mathcal{U}[280, 310]$\\3s: $\mathcal{U}[210, 230]$; 4s: $\mathcal{U}[240, 270]$} &\makecell[c]{thickness: $\mathcal{U}[1.2, 1.4]$\\2: $\mathcal{U}[14, 16]$}\\
    \midrule
    5 &  \makecell[c]{1: $\mathcal{U}[2.5,3]$\\3: $\mathcal{U}[1.3, 1.5]$} & \makecell[c]{1s: $\mathcal{U}[95, 110]$; 1e: $\mathcal{U}[300, 310]$\\3s: $\mathcal{U}[210, 230]$} &\makecell[c]{thickness: $\mathcal{U}[1, 1.2]$\\2: $\mathcal{U}[14, 16]$; 4: $\mathcal{U}[1, 2]$}\\
    \bottomrule
    \end{tabular}
    \vspace{15pt}
    \label{random_distributions}
\end{table*}

\begin{itemize}
    \item \textbf{3D Workpiece}. The workpieces are determined by the cross-sections and characteristic lines. The cross-sections will affect the overall structural force distribution. For practical concerns, we select 5 representative types of cross-section structures, as depicted in Figure~\ref{data_cross_section_appendix}, from practical observation. These cross-sections can cover most practical products in term of the topology structure and size. Each kind of cross-section has different number, radian and radius of arcs, various thicknesses and heights. In each kind of cross-section, the radius of arcs, the height and thickness are also sampled from specific distributions. The concrete configurations are listed in Table~\ref{data_cross_section_distribution}. For each type of cross-section structure, we generate 600 different samples. For the characteristic line of the workpiece, we place the initial straight workpiece on the x-axis from original point to the maximal length and randomly sample the length from a uniform distribution $\mathcal{U}[505, 550]$ (unit: mm). 
    
    \item \textbf{3D Mold}. In order to ensure the workpiece can contact the mold tightly during the deformation, the characteristic lines for the mold should be smooth and convex. We generate two 1/4 elliptical arcs separately on two perpendicular 2D planes, then combine them into a 3D curve in space. These different elliptical arcs are determined by different ellipse parameters with uniform distributions for varying curvatures. Consider a 2D elliptic formula $\frac{x^2}{a^2}+\frac{y^2}{b^2}=1(a>b>0)$ and let $(c, 0)$ be its focus point, we use three uniform distributions to control the generation, which are: $\frac{c}{a}\sim\mathcal{U}[0.1, 0.3]$, $\frac{b}{a}\sim\mathcal{U}[0.1, 0.3]$ and $a\sim\mathcal{U}[700, 900]$ (unit: mm).
    
    \item \textbf{Motion parameters}. The motion parameters consist of 6 degrees of freedom. including spatial displacement $(u_x,u_y,u_z)$, and the rotations $(r_x, r_y, r_z)$. They are calculated by a classical involute approach \cite{arnold2012classification} based on the characteristic line of the mold. 
\end{itemize}

With all the 3000 sets of data, Abaqus \cite{khennane2013introduction, abaqus2011abaqus}, a software based on FEM, is applied to perform the computations. The mold is set to be rigid, and the workpiece is set as elastic-plastic aluminum. In Abaqus setup, the explicit dynamics solver \cite{barbero2023finite} is adopted for simulation calculation and the implicit static solver \cite{barbero2023finite} is for the rebound calculations.

\section{Two-Loop Mold Design Paradigm}
\label{Two-Loop Mold Design Paradigm}

\begin{figure*}[b]
    \centering
    \includegraphics[width=0.9\linewidth,keepaspectratio]{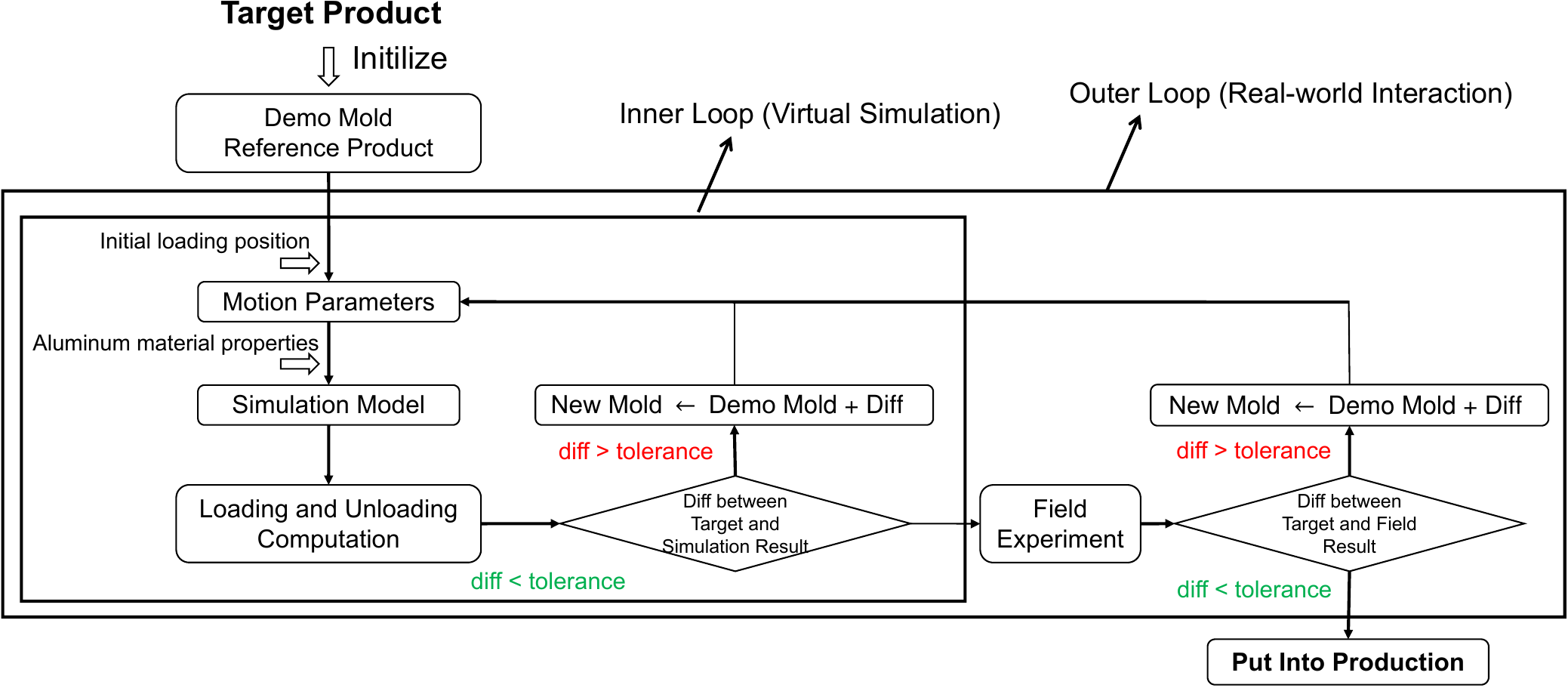}
    \caption{Two-loop mold design paradigm.}
    \label{two-loop}
\end{figure*}

Errors in mold design process are mainly categorized into simulation error (SE), application error (AE) and production error (PE). The SE is the discrepancy between the result of simulation and the target shape, defined within the virtual space. However, there is often a gap between simulation results and their practical application due to various sources of factors arise at different stages of process. For example, mold production inherently involves certain tolerances which are introduced by the manufacturing machines. Installation tolerance arises when installing the mold onto the machine. The motion parameters of working arms may exhibit zero-point drift, and the properties of workpiece change over time, among other factors. The AE is defined as the error generated when simulation results are directly implemented in production, within 10mm. This gap is inevitable due to the inherent complexities. Then on-site adjustments will be carried out. On-site adjustment methods include machining of the mold by tools directly, as well as fine-tuning the parameters of working arms. By employing these methods, the final PE will be kept below 1mm for production. 

In proposed two-loop mold design paradigm in Figure~\ref{two-loop}, we use inner and outer loop to guarantee the reliability and meet all errors mentioned above. In inner loop, we use simulation and displacement compensation method \cite{cafuta2012enhanced} to iteratively obtain a simulation result meet the SE and AE. The simulation is FEM in traditional and LaDEEP now. Then through on-site adjustment, PE could be met in most cases. If, after several adjustments, the on-site results still fail to meet production standard, it is necessary to feedback the actual outcomes to the simulation system. This feedback process, called outer loop, is crucial as it is a recalibration for the compensation. And it ensures a seamless transition from the virtual simulation space to the real-world production environment.

We start from a naïve design of mold (say a straight one), simulate the deformation process, compute the distance between the simulation results and the desired shape. This distance helps us to re-design the mold, and the computations start over. This cycle is conducted purely in computations, and terminates until the workpiece deforms as designed after rebound. The mold will be delivered to the factory and produced in real world. There will also be errors after real-world production, as there are errors on simulation, which will be feedback to LaDEEP for further design.

\end{document}